\documentclass[lettersize,journal]{IEEEtran}
\usepackage{amsmath,amsfonts}
\usepackage{algorithmic}
\usepackage{algorithm}
\usepackage{array}
\usepackage[caption=false,font=normalsize,labelfont=sf,textfont=sf]{subfig}
\usepackage{textcomp}
\usepackage{stfloats}
\usepackage{url}
\usepackage{verbatim}
\usepackage{graphicx}
\usepackage{cite}
\usepackage{hyperref}
\usepackage{bm}
\begin{document}

\title{From Flies to Robots: \\ Inverted Landing in Small Quadcopters with Dynamic Perching}

\author{Bryan Habas$^{1}$, 
Bo Cheng$^{1}$, \textit{IEEE, Member}% <-this % stops a space
\thanks{$^{1}$Biological and Robotic Intelligent Fluid Locomotion Lab, Department of Mechanical Engineering, The Pennsylvania State University, University Park, PA 16802, USA. Corresponding to B.C. {\tt\small buc10@psu.edu}}
\thanks{Manuscript submitted for review February 29, 2024}}
% <-this % stops a space

% The paper headers
% \markboth{IEEE Transactions on Robotics, 2024}%
% {Shell \MakeLowercase{\textit{et al.}}: A Sample Article Using IEEEtran.cls for IEEE Journals}

% \IEEEpubid{0000--0000/00\$00.00~\copyright~2021 IEEE}
% Remember, if you use this you must call \IEEEpubidadjcol in the second
% column for its text to clear the IEEEpubid mark.

\maketitle

\begin{abstract}
Inverted landing is a routine behavior among a number of animal fliers. However, mastering this feat poses a considerable challenge for robotic fliers, especially to perform dynamic perching with rapid body rotations (or flips) and landing against gravity. Inverted landing in flies have suggested that optical flow senses are closely linked to the precise triggering and control of body flips that lead to a variety of successful landing behaviors. Building upon this knowledge, we aimed to replicate the flies' landing behaviors in small quadcopters by developing a control policy general to arbitrary ceiling-approach conditions. First, we employed reinforcement learning in simulation to optimize discrete sensory-motor pairs across a broad spectrum of ceiling-approach velocities and directions. Next, we converted the sensory-motor pairs to a two-stage control policy in a continuous augmented-optical flow space. The control policy consists of a first-stage Flip-Trigger Policy, which employs a one-class support vector machine, and a second-stage Flip-Action Policy, implemented as a feed-forward neural network. To transfer the inverted-landing policy to physical systems, we utilized domain randomization and system identification techniques for a zero-shot sim-to-real transfer. As a result, we successfully achieved a range of robust inverted-landing behaviors in small quadcopters, emulating those observed in flies.  
\end{abstract}

\begin{IEEEkeywords}
Aerial Systems: Mechanics and Control, Biologically-Inspired Robots, Learning and Adaptive Systems, Aerial Systems: Applications
\end{IEEEkeywords}

\section*{SUPPLEMENTARY MATERIAL}
\noindent \textbf{Video:} \url{https://youtu.be/MT0rrtnQ0ME} \\

\section{Introduction}
\IEEEPARstart{P}{erching} is a feat routinely performed by animal fliers with robustness and accuracy, as seen in birds\cite{lee1993visual, roderick2019birds,carruthers2010mechanics}, bees \cite{srinivasan2000honeybees,baird2013universal,tichit2020accelerateda,tichit2020acceleratedb,goyal2022bumblebees,goyal2019flight}, flies\cite{liu2019flies,balebail2019landing,borst1990flies},  and bats\cite{lee1995steering,riskin2009bats}. With the ability to perch on surfaces that are arbitrarily oriented or moving unpredictably, animal fliers can land in unstructured environments to surveil territory, hitchhike on larger animals for travel to new locations, pollinate plants, or rest.

Achieving similar perching abilities in flying robots is crucial for enabling their fully autonomous operation in unstructured environments \cite{zufferey2008bio}, with the potential for enabling or augmenting  long-term inspections, surveillance, reconnaissance, and the rapid release and retrieval of robots for various missions\cite{restas2015drone, mishra2020drone,kim2018drone,seo2018drone,irizarry2012usability}. A major drawback of these robot systems over recent years has been their limited battery life, which typically only sustain maximum flight times on the scale of tens of minutes \cite{cesare2015multi}. For this reason, the ability to perch on targeted objects can greatly expand the operational lifetime of quadrotor robots. This capability could be particularly applicable in urban environments, where flat surfaces like walls and ceilings are abundant and many tasks such as camera surveillance, sensor reading, and acting as a radio relay in disaster zones do not require continuous flight \cite{thomas2016aggressive,mishra2020drone,restas2015drone}.

\begin{figure}[!t]
    \centering
    \includegraphics[width=1.0\columnwidth]{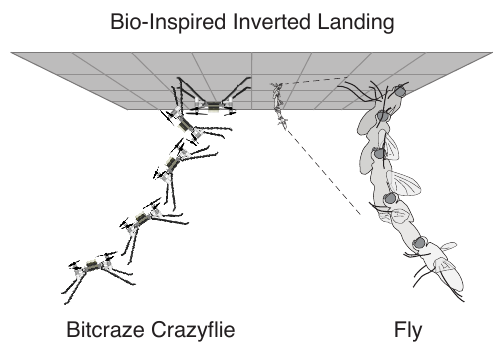}
    \caption{Bio-inspired inverted landing. An illustration comparing an example of inverted landing in  a small quadcopter and a blue-bottle fly\cite{liu2019flies}. Both an at-scale and a scaled-up version of the landing sequence of the blue-bottle fly are shown.}
    \label{fig:Intro_Figure}
\end{figure}

\subsection{Dynamic Perching and Challenges}

\noindent Despite the impressive aerial agility and load-carrying capacity exhibited by small aerial robots like quadrotors, they have yet to achieve the perching capabilities observed in animal fliers. Indeed, \emph{dynamic perching}—characterized by rapid maneuvers and subject to velocity constraints—still poses one of the most significant challenges for aerial robots. This skill becomes vital when landing on rapidly moving or inverted (ceiling-like) surfaces, necessitating the flier to accelerate against gravity and rapidly alter its orientation. In such scenarios, fliers must solve a complex distance-velocity-attitude sensing and control problem under stringent time and motor constraints. This requires the small flier to control its 3D linear velocity\cite{Kendoul2014,baird2013universal}, accurately predict imminent impact\cite{van2012visual,habas2022optimal}, and timely adjust its body orientation and landing gear/legs relative to the perching surface through rapid angular maneuvers. These maneuvers must be executed within milliseconds to ensure a  touchdown process with proper body inversion and contact force \cite{liu2019flies}.

Challenges associated with perching extend beyond the maneuver itself, involving the limited computational resources available for sensing, control, and planning in small fliers\cite{habas2022optimal,mao2021aggressive}. Achieving accurate onboard sensory estimations within the stringent time constraints of dynamic perching is a formidable task (e.g., a delay of just a few milliseconds could result in a crash), especially in unstructured environments. Aerial robots are also commonly under-actuated, which makes it difficult to execute aggressive maneuvers demanding uncoupled linear and angular acceleration. Moreover, aggressive acrobatic flight necessitates high thrust and extreme angular accelerations that can push the motor limits of the robots, potentially leading to unmodeled behaviors. Therefore, a successful perching strategy must consider these limitations and rely on computationally efficient  algorithms in order to replicate the perching behaviors of animal fliers. Importantly, physically-embodied intelligence\cite{Kovac2016,pfeifer2007self}, seen in the landing gear or legs, can help to mitigate computational demands and bolster robustness.

Inverted landing (Figure \ref{fig:Intro_Figure}) also presents a significantly greater challenge than landing on horizontal or vertical surfaces due to the fact that gravity works against the robot as it approaches the landing surface. Furthermore, to truly emulate the behavior of flies, the robots must be capable of approaching the surface from a variety of directions \cite{liu2019flies}. This introduces stringent and variable constraints on both the linear velocity—essential to prevent downward falls and ensure sufficient adhesion upon contact—and the body orientation, which must be accurately controlled to ensure proper contact and avoid collision.

\subsection{Inverted Landing in Flies}

\noindent Flies, for example, have successfully solved the dynamic perching problem \cite{hyzer1962flight}. Their solution is a sequence of well-coordinated maneuvers, completed in less than 100 ms, that facilitate an inverted landing\cite{liu2019flies}. To initiate the process, they engage in an upward acceleration, which is followed by a rapid rotation of their body and an extension of their legs; the sequence concludes with the flies executing a leg-assisted body swing, using their fore-legs—now firmly attached to the ceiling—as a pivot point. Moreover, it has been shown that different ceiling-approach directions have led to diverse landing behaviors in terms of axes of rotation and angular rates of body maneuvers\cite{liu2019flies}. More importantly, the success of these landing can be explained by the timely triggering and proper control of body rotational maneuvers (or flips) that are both strongly correlated to a number of optical flows that flies' visual system can extract.  

The landing process in flies involves a combination of mechanical and computational intelligence as they rapidly and precisely position their legs and tarsi prior to impact (Figure \ref{fig:Intro_Figure}), which are then used to dissipate impact force or even recover from a failed landing. Known for their efficient visuomotor mapping, flies possess direct connections between their visual system and motor neurons. These connections facilitate rapid motor program selection and timing, enabling reactive aerobatic maneuvers \cite{zufferey2008bio}. In particular, the insect visual system is evolved for efficient extraction of optical flow\cite{borst2009drosophila}, which encodes translational velocity, distance to a surface, and time-to-contact without the need for arduous feature identification and tracking. Visual cues like optical flow are widely used in nature and provide a computationally efficient and effective way for animals of various sizes to follow robust perception-based landing trajectories \cite{srinivasan2000honeybees,lee1993visual}. 

\subsection{Goal and Contributions of the Current Work}

\noindent Building upon our previous work \cite{habas2022optimal}, which focused primarily on identifying a capable three-dimensional control policy region and exploring the effects of landing gear design on landing success through initial simulation and limited experimental testing, this journal submission significantly advances our research. Our current study introduces a general two-stage control policy inspired by the inverted landing strategy of flies that centers on the mappings from optical flow-based cues to the triggering and control of rotational maneuvers. This control policy generalizes the discrete sensory-motor pairs obtained using reinforcement learning, utilizing framework shown in our conference proceedings, inside our further refined simulation, and efficiently maps an augmented optical-flow space to motor control actions. Our strategy—unlike existing methods reliant on trajectory planning and tracking—promotes the emergence of diverse behaviors via motor control and touchdown processes, thus enabling the close integration of computational and mechanical intelligence.

To validate our approach, we transferred the generalized inverted-landing policy from simulation to real-world experiments with a physical quadrotor, employing domain randomization and system identification techniques for zero-shot transfer. Finally, through a series of experiments under a variety of approach conditions, we demonstrated a range of successful inverted landings. Additionally, we investigated the influence of leg design geometry on the overall success and robustness of the inverted landing.

\subsection{Organization of the Paper}
\noindent In Section II, we delve into the literature on robotic landing, exploring both their computational and mechanical intelligence-based methods. The proposed two-stage general control policy for inverted landing, and its training methodology, are presented in detail in Section III. Section IV elaborates on our simulation setup and the experimental testing process. In Section V, we discuss the performance outcomes of our work, the results achieved, and the potential benefits of varying landing gear designs. The final section, Section VI, draws conclusions from our study and suggests avenues for future research. Additional details on system identifications and relevant explanations are provided in Appendix A.

\section{Existing Work on Robot Landing}

\subsection{Computational Intelligence} 
\noindent In the literature of robot perching, a significant number of methods are based on the generation and optimization of target landing trajectories that the robots can follow. These methods offer advantages such as power efficiency, customizable boundary conditions, and the ability to execute aggressive maneuvers \cite{thomas2016aggressive,mao2022robust}. However, they often rely heavily on the provision of real-time external positioning data to the robot or the use of an off-board computer for computing optimal trajectories \cite{hang2019perching,chi2014optimized,thomas2016aggressive}, rendering them less than ideal for rapid dynamic perching and exploring the mechanical intelligence for emergent landing behaviors.

Many robotic perching studies have explored using onboard cameras to compute flight trajectories, such as through state estimation algorithms, to land on vertical surfaces \cite{mao2022robust,dougherty2016monocular,alkowatly2015bioinspired,yang2013onboard}. While these methods offer advantages over external positioning systems, they tend to experience tracking degradation during aggressive maneuvers and require feature tracking of a predefined landing target, making them less practical in unstructured environments \cite{mao2022robust}. In contrast, featureless visual-based landing strategies rely on calculating optical flow-based sensor values from monocular cameras, without the need for feature tracking, enabling them to be applied in a wider range of outdoor settings \cite{alkowatly2015bioinspired,das2018bio,luo2017vision,kendoul2014four}. Despite their simplicity and computational efficiency, there is still ongoing research into how to optimize these methods for robustness and adaptability to varying environmental conditions.

\subsection{Mechanical Intelligence}
\noindent Within the realm of robotic landing research, the design of the physical system plays a pivotal role in enabling robots to attach to a variety of landing surfaces and objects through their mechanical intelligence. This has resulted in a broad range of approaches, with many inspired by nature. These popular methods frequently involve the use of grasping mechanisms that allow the robot to hang or perch on branch-like objects \cite{chi2014optimized,thomas2013avian,luo2017vision,thomas2016visual,yu2022implementation,bai2021design,doyle2012avian,hsiao2022mechanically,zhang2020compliant}. Other approaches include the utilization of dry-adhesive materials that mimic gecko skin, enabling the robot to attach to any orientation of exceptionally smooth surfaces \cite{kalantari2015autonomous,thomas2016aggressive,he2020optimized}, as well as novel devices such as suction cups \cite{hsiao2022novel,wopereis2016mechanism}, barbed hooks \cite{pope2016multimodal}, and magnets \cite{huang2021biomimetic}. The variety of attachment devices offers unique advantages and benefits, making them well-suited for different landing scenarios.

\begin{figure*}[!t]
    \centering
    \includegraphics[width=1.0\textwidth ]{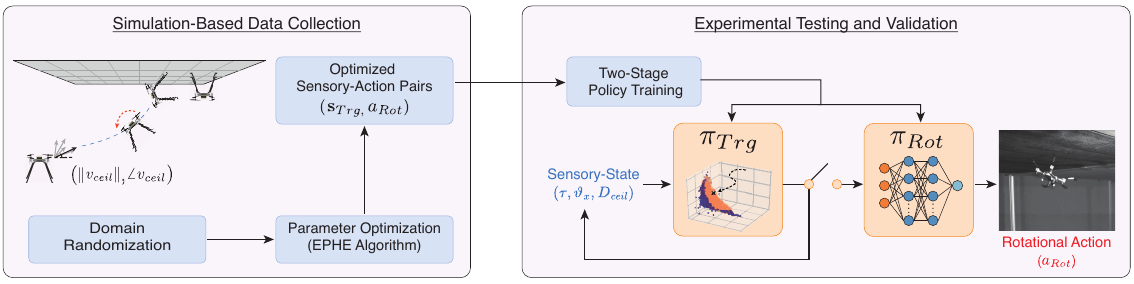}
    \caption{Block diagram illustrating the data collection process through simulation, the training of our two-stage policy using the collected data, and the implementation of the policy in experimental tests.}
    \label{fig:Block_Diagram}
\end{figure*}

Although there is a growing variety of landing devices in robotics, research analyzing how the geometry of these devices affects landing robustness and landing strategy is still limited. Some authors have partially investigated these effects, such as by using active skids to land on inclined surfaces \cite{kim2021autonomous} or by modeling the material properties of landing gear \cite{huang2021biomimetic,ni2022research,dunlop2021modeling}, but comprehensive research in this area is still required. Our work contributes to this area by analyzing the effects of varying geometries on highly dynamic landings and impacts for inverted landing scenarios. While our analysis framework and methodology are specifically tailored to this case, it can be easily adapted to evaluate the effects of mechanical intelligence from different landing gear designs on surfaces of varying orientations, such as vertical, horizontal, or inclined.

Several remarkable successes have been recorded in the field of robotic perching on inclined\cite{mao2021aggressive,thomas2016aggressive} or moving surfaces\cite{liu2022hitchhiker,baca2019autonomous,ZhangIEEE2021,paris2020dynamic}. Such methods typically concentrate on the planning, optimization, and tracking of dynamically-feasible trajectories to bring a robot within some empirically constrained states before touchdown \cite{mao2021aggressive,paneque2022perception,liu2022hitchhiker}. Despite their successes, these methods pose significant computational challenges when applied to rapid dynamic perching in small robots and are not ideally equipped to leverage mechanical intelligence during the touchdown process. Another limitation lies in their reliance on empirically-tuned constraints on terminal states that are specific to a particular type of landing, as well as on computationally-intensive state estimation of both the robot and the perching target, often facilitated by external motion tracking systems. Additionally, traditional optical-flow-based methods like tau-theory \cite{lee1993visual}, which rely on zero-velocity contact, are inadequate for fostering diverse landing behaviors on inverted surfaces.

\section{Method: Two-stage, General Control Policy for Inverted-Landing}

\subsection{Overview}
\noindent The training of our generalized control policy for inverted landings begins with simulating various ceiling-approach conditions; similar to those observed in flies (see Methods: Simulation and Experimental Setup). Using model-free reinforcement learning, we collected discrete sensory-motor pairs, specifically triggering sensory state (\mbox{$\textbf{s}_{Trg}$}) within the augmented optical-flow space ($\mathbb{OF}_{a}$) and motor action for rotational maneuvers (\mbox{$a_{Rot}$}), which led to optimal motor control to maximize the success rate of inverted landings under each ceiling-approach condition \cite{habas2022optimal}. 

To transform the discrete sensory-motor pairs into a generalized policy, we developed a two-stage control policy. In the first stage, we identified a cluster within $\mathbb{OF}_{a}$ where the discrete sensory states $\textbf{s}_{Trg}$ linked with successful landings (exceeding a defined threshold of success rate). We used this cluster to construct a continuous boundary function around the region, differentiating optimal triggering states from sub-optimal or failed ones, giving rise to a \emph{Flip Trigger Policy} ($\pi_{Trg}$). This policy allowed the robot agent to detect when inside the optimal sensory region to initiate body angular maneuvers (or flips). To model the $\pi_{Trg}$ (or the boundary region of $\textbf{s}_{Trg}$ in $\mathbb{OF}_{a}$), we used a One-Class Support Vector Machine (OC-SVM). In stage two, we utilized supervised learning to train a simple neural network that forms the \emph{Flip Action Policy} $\pi_{Rot}$, a continuous sensory-motor map that generalizes discrete sensory-motor pairs. $\pi_{Rot}$ provides the body-moment command, controlling the rotational maneuver in a feedforward fashion. Thus, under this control policy, if the robot identifies its location within the $\mathbb{OF}_{a}$ region specified by the optimized $\pi_{Trg}$ policy, it generates a body rotational moment according to $\pi_{Rot}$ to execute an inverted landing. An overview of our two-stage policy can be seen in Figure \ref{fig:Block_Diagram}.

\begin{figure}[!t]
    \centering
    \includegraphics[width=1.0\columnwidth]{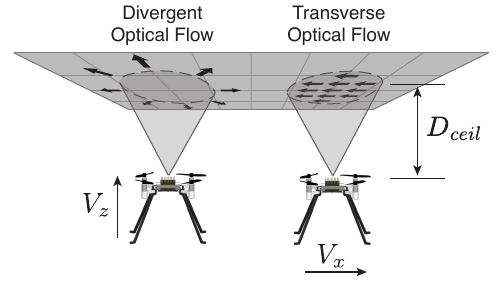}
    \caption{(a) Diagram illustrating divergent optical flow: as the quadrotor nears the ceiling, observed points radiate outward. (b) Diagram showcasing transverse optical flow, where feature points move horizontally across the field of view during the robot's translation beneath the ceiling.}
    \label{fig:Optical_Flow_Visual}
\end{figure}

\subsection{Augmented-Optical Flow Space}

\noindent The sensory states $\textbf{s}_{Trg} \in \mathbb{OF}_{a} \subseteq \mathbb{R}^{3}$ used as inputs for our two-stage control policy comprise of two visual sensory cues, including time-to-contact ($\tau$) and a transverse optical flow term ($\vartheta_x$). These are augmented with a robot-to-surface distance metric ($D_{ceil}$) \cite{habas2022optimal}. The variable $\tau$ encodes the time until the robot reaches a surface, assuming a constant velocity and heading. Studies show that in flies and pigeons, $\tau$ provides predictive information about when to initiate a rapid angular maneuver, converting a potential collision into a successful perch \cite{liu2019flies,lee1993visual}. As a visual cue, $\tau$ is the mathematical inverse of divergent optical flow, or the Relative Retinal Expansion Velocity $\left( RREV \right)$, which represents how rapidly objects expand within the camera's Field of View (FoV) as the robot approaches an object (see Figure \ref{fig:Optical_Flow_Visual}a). $\tau$ has been shown to directly encode the environment's affordance \cite{gibson2014ecological} (i.e., comprehending the robot's potential dynamic fit within the environment or perception-action \cite{Kendoul2014}). Furthermore, various works support the utilization of this term in $\tau$-theory, where the time derivative remains constant within the animal system, and a single parameter value determines the entire landing trajectory \cite{lee1993visual}.

While $\tau$ can be derived directly from onboard visual inputs without the need to measure the relative position or velocity between the robot and the landing target, in this study, it was emulated using external motion tracking data as per the following equation:

\begin{equation}
    \label{Eq:Tau_Eq}
    \tau = \frac{D_{ceil}}{V_{z}}.
\end{equation}

Where $D_{ceil}$ represents the distance to the ceiling, and $V_{z}$ denotes the vertical velocity of the system.

Moreover, the transverse optical flow term, denoted as $\left( \vartheta_x \right)$, encodes the relative distance and horizontal velocity of the robot with respect to the landing surface. This term is determined by both the horizontal velocity of the robot ($V_x$) and its distance relative to the landing surface ($D_{ceil}$), refer to Figure \ref{fig:Optical_Flow_Visual}b\cite{srinivasan2000honeybees,chahl2004landing}. $\vartheta_x$ is anticipated to correlate directly with the robustness of the robot's landing and the resulting body-moment required for the flip action. The value of $\vartheta_x$ can be emulated using external motion tracking data through the following equation:

\begin{equation}
    \label{Eq:Theta_x_Eq}
    \vartheta_x = \frac{V_{x}}{D_{ceil}}.
\end{equation}

Finally, the authors' previous research indicated that the reliance on purely the two aforementioned optical flow terms was insufficient to fully define the policy-space \cite{habas2022optimal}. Specifically, the $\textbf{s}_{Trg}$ regions corresponding to successful and failed landings couldn't be separated within the optical flow space $\mathbb{OF} ( \tau, \vartheta_x) \subseteq \mathbb{R}^{2}$. To overcome this obstacle, it was necessary to augment the optical flow space with an additional parameter such as $D_{ceil}$ or the robot's airspeed. Although insects can sense their airspeed through their antennae, it is challenging for small robotic fliers to achieve the same. Hence, we chose $D_{ceil}$ as the augmentation parameter. This measurement can be potentially derived using a laser distance sensor or estimated by fusing the system's onboard accelerometer measurements with the time derivative of $\tau$ \cite{chirarattananon2018direct}. Consequently, the triggering sensory state $\textbf{s}_{Trg} \in \mathbb{OF}_{a} \left(\tau,\vartheta_x,D_{ceil}\right) \subseteq \mathbb{R}^{3}$ forms the input tuple for our generalized control policy.

Notably, although the optical flow terms were emulated in real-time using a Vicon motion capture system in lieu of a physical sensor or image processing algorithm, several works and methods currently exist for efficiently calculating these values from a simple monocular camera sensor with featureless-based optical flow algorithms \cite{horn1981determining,horn2009hierarchical,chirarattananon2018direct}. In future work, we plan to directly estimate these sensor-derived terms and incorporate them into our landing strategy to achieve fully onboard inverted landing.

\subsection{Reinforcement Learning-Based Optimization of Sensory-Action Pairs for Variable Ceiling Approach Conditions}

\noindent Instead of directly obtaining a control policy applicable to any arbitrary ceiling-approach condition, we determined the optimal sensory-motor pairs $(\textbf{s}_{Trg},a_{Rot})$ individually for each simulated ceiling-approach condition. For each learning trial, characterized by a constant velocity magnitude $\mathbf\|{v}_{ceil}\|$ and approach angle $\mathbf\angle{v}_{ceil}$, we used policy gradient Reinforcement Learning (RL) to optimize the optimal trigger timing threshold, $\tau_{cr}$, and a corresponding body-rotation motor-action, $a_{Rot}$, that maximized the reward function related to successful landings (as detailed in the following section). This body-rotation action was applied by adjusting the fore/aft motor thrusts to induce an angular moment about the robot's pitch axis. The learning process was repeated three times for each $V \in \left[1.5 - 3.5 \right]$ m/s and $\phi \in \left[30^{\circ} - 90^{\circ} \right]$, incremented by $0.1$ m/s and $3.75^{\circ}$, respectively. We recorded the complete set of optimized $(\textbf{s}_{Trg},{a_{Rot}})$ pairs for all approach conditions, as well as the corresponding landing success rates.

In this optimization process, our RL approach utilized the EPHE policy-gradient-based algorithm\cite{wang2016based}. We chose this algorithm for its fast convergence and adaptive learning rate, which reduces the need for extensive hyperparameter tuning. The algorithm and its derivation are detailed in the work by Wang et al. \cite{wang2016based}. In our application, the EPHE algorithm determines an optimal set of parameters by generating a series of parameterized Gaussian distributions $\bm{O} = \left[O_{\tau_{cr}}, O_{Rot}\right]$ with $\bm{O} = \mathcal{N} (\bm{\mu}, I\bm{\sigma^2})$, defined by the vectors $\bm{\mu} = \left[\mu_{\tau_{cr}}, \mu_{Rot} \right]$ and $\bm{\sigma} = \left[\sigma_{\tau_{cr}}, \sigma_{Rot}\right]$. These vectors are then optimized through interaction with the environment until the distributions approximate a deterministic value.

\begin{figure}[!t]
    \centering
    \includegraphics[width=1.0\columnwidth]{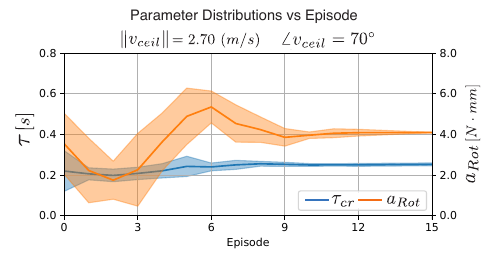}
    \caption{The figure illustrates the convergence of $\tau_{cr}$ and $a_{Rot}$ values obtained through reinforcement learning in simulation. These values are used to establish an optimal sensory-motor pair. The shaded regions indicate a distribution range of $\mu_i\pm 2\sigma_i$ for samples taken in each episode.}
    \label{fig:EPHE_Convergence_Fig}
\end{figure}

The detailed RL process is as follows: Each landing trial began with the quadrotor in an initial hovering state, after which it was programmed to follow a constant velocity trajectory with a specified speed $V$ and flight angle $\phi$ that leads to collision with the ceiling surface. This velocity trajectory was maintained throughout the landing trial. Following this, the agent performed several rollouts where, for each rollout, a policy $\bm{\theta} = \left[ \tau_{cr}, a_{Rot} \right]$ was sampled from the current distribution $\bm{O}$ and used to execute a landing maneuver. To do so, the quadrotor observed the current $\tau$ value until it exceeded the $\tau_{cr}$ threshold set by the sampled policy, at which point it executed the body flip motor-action $a_{Rot}$ dictated by the same policy parameters. The rollout terminated based on whether the landing was successful or a timeout threshold was reached. At this juncture, a reward for the rollout was calculated before the process was repeated for a new rollout and sampled policy. After N rollouts were performed, the best K rewards and policy parameters were used to update the original policy distribution $\bm{O}$; this constituted a complete episode. This process was repeated for 15 episodes or until there was convergence to a nearly deterministic set of optimized parameters which maximized the probability of a successful landing for a given set of velocity conditions. In each learning trial, convergence to a deterministic policy typically occurred within 100 rollouts. Figure \ref{fig:EPHE_Convergence_Fig} provides an example of such convergence, illustrating the convergence behavior for a landing trial with a velocity magnitude of 2.70 m/s and a flight angle of 70 degrees relative to the horizon.

\subsection{Design of Reward Function}

\noindent In the domain of reinforcement learning, constructing an effective reward function is crucial for successful learning, swift convergence, and the integrity of the agent's learned policy. Applying the principles of curriculum learning to our reward function design allowed for consistent and reliable policy convergence. This method incrementally extends the overall reward attainable by the system with each learned skill in the inverted landing procedure, thereby guiding the system towards mastering more complex behaviors. These skills and their associated reward functions were characterized to: minimize the quadrotor's distance to the ceiling as depicted in (\ref{Eq:r_d}); initiate the flip maneuver prior to surface contact as per (\ref{Eq:r_tau}); adjust the impact angle to ensure initial contact is made by the fore-legs as depicted in (\ref{Eq:r_theta}); and fine-tune the preceding conditions to guarantee the achievement of reliable inverted landing as demonstrated in (\ref{Eq:r_legs}) \cite{habas2022deep}.

The constants ($c_0,c_1$), present in equations (\ref{Eq:r_d}) and (\ref{Eq:r_tau}), serve to normalize their corresponding reward functions and adjust the breadth of their clipped zones. By tuning these elements, the reward functions can direct the system to converge within a practical array of values, avoiding an overly specific fixation on a single value. This principle is notably applied in (\ref{Eq:r_tau}), which advocates for the flip maneuver to fall within the ideal trigger interval $\tau_{cr} \in [0.15,0.25]$, without overly dictating the timing of the maneuver \cite{habas2022optimal}.

A penalty factor of $r_{legs} \leftarrow r_{legs}/3$ modifies equation (\ref{Eq:r_legs}) in instances of propeller or body contact. This adjustment serves to deter any resulting policy that might induce structural harm to the quadrotor. Consequently, to compute the total reward for the entire episode, the individual rewards were proportionally weighted and summed accordingly, $r = 0.05\cdot r_{d} + 0.1\cdot r_{\tau} + 0.2\cdot r_{\theta} + 0.65\cdot r_{legs}$.

\begin{equation}
    \label{Eq:r_d}
    r_{d} = clip\left(\frac{1}{|d_{min}|},0,c_0\right) \cdot \frac{1}{c_0},
\end{equation}

\begin{equation}
    \label{Eq:r_tau}
    r_{\tau} = clip\left(\frac{1}{|\tau_{trg}-0.2|},0,c_1\right) \cdot \frac{1}{c_1},
\end{equation}

\begin{equation}
    \label{Eq:r_theta}
    r_{\theta} = 
            \begin{cases} 
            \frac{|\theta_{impact}|}{120^{\circ}} & 0^{\circ} \le |\theta_{impact}| < 120^{\circ} \\
            1.0 & 120^{\circ} \le |\theta_{impact}| \le 180^{\circ} 
           
            \end{cases},
\end{equation}

\begin{equation}
    \label{Eq:r_legs}
    r_{legs} = 
            \begin{cases} 
            1.0 & N_{legs} = 3 \ || \ 4 \\
            0.5 & N_{legs} = 1 \ || \ 2 \\
            0 & N_{legs} = 0 \\
            
            \end{cases}.
\end{equation}

\subsection{Develop Two-Stage, General Control Policy in Continuous Domain}

\noindent  Next we transformed the optimized $(\textbf{s}_{Trg},{a_{Rot}})$ pairs  to a  general control policy  within a continuous domain, applicable to a wider range of approach scenarios. This generalized policy first identified the bounds of the triggering state region, i.e. the region of $\textbf{s}_{Trg} \in \mathbb{OF}_{a}\left(\tau,\vartheta_x,D_{ceil}\right)$ for which triggering the flip maneuver can lead to a successful landing. Subsequently, it generated the optimal rotation action based on the specific triggering state $\textbf{s}_{Trg}$. This two-stage approach leads to our control policy consisting of the \emph{Flip Trigger Policy} ($\pi_{Trg}$) and the \emph{Flip Action Policy} ($\pi_{Rot}$). In crafting this policy, we employed a combination of unsupervised and supervised machine learning methods. Notably, the current approach exemplifies a grey-box approach, allows for direct identification of policy region for triggering, emergent policy behaviors, and landing success rate, therefore allowing us to examine the robustness of landing. This trait is also essential for future work using trajectory planning to fly the robots into this policy region for successful landing.
This stands in contrast to the conventional black-box methods utilized in Deep RL algorithms, which directly generate both the triggering and rotation actions for inverted landing without insight to its inner workings \cite{habas2022deep}.

\subsubsection{Flip Trigger Policy (One-Class SVM)} 

Since every $(\textbf{s}_{Trg},{a_{Rot}})$ pair corresponds to a specific success rate, we can identify this optimized cluster by setting a success rate threshold of $80\%$. Consequently, the \emph{Flip Trigger Policy} can be articulated as a closed boundary function in the $\mathbb{OF}_{a}$ space, indicative of this specific cluster. We then define ${\pi}_{Trg}(\textbf{s}_{Trg})$ to output a binary value when the current sensory-state lies within this region, triggering the flip maneuver accordingly.

Owing to the highly nonlinear nature of this region, we employed unsupervised machine learning, particularly outlier detection algorithms, to shape this boundary function. These algorithms are a subset of multi-class clustering techniques, focusing on separating the data into a single class and evaluating whether new data lies inside or outside the established group. Noteworthy algorithms suitable for this application include Isolation Forest \cite{liu2008isolation}, Local Outlier Factor \cite{cheng2019outlier}, Robust Principal Component Analysis \cite{breunig2000lof}, and One-Class Support Vector Machines \cite{scholkopf1999support,ma2003time,amer2013enhancing}. In this context, we opted for utilizing a One-Class Support Vector Machine (OC-SVM) due to its superior efficiency in decision-making calculations following the initial training.

\begin{figure}[!t]
    \centering
    \includegraphics[width=1.0\columnwidth]{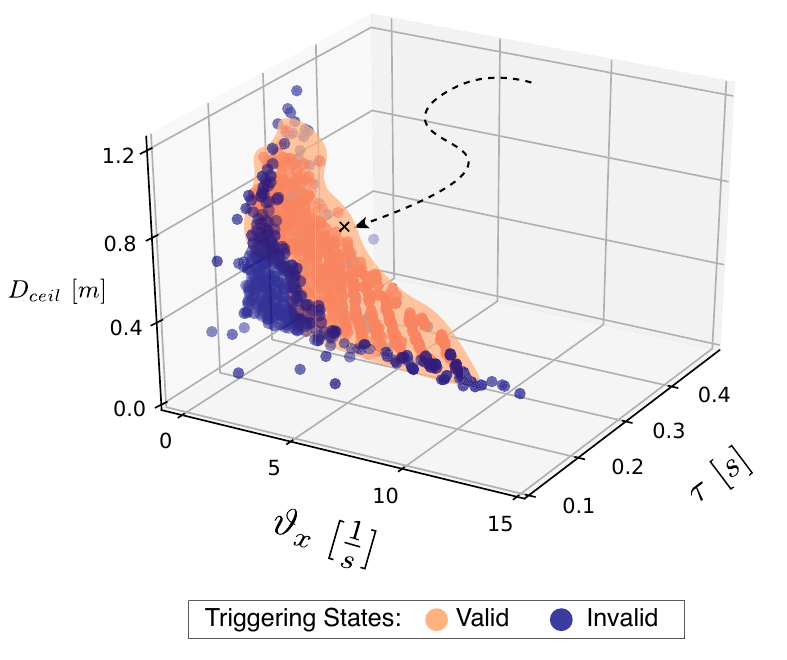}
    \caption{The decision boundary formed by the One-Class SVM, implemented for the \emph{Flip Trigger Policy}, is shown. This boundary distinguishes between valid and invalid triggering states for inverted landings, consistent with the policy ${\pi}_{Trg}(\textbf{s}_{Trg})$. A representative trajectory is also depicted, indicating the exact timing of the rotational maneuver trigger upon intersecting with the boundary region.}
    \label{fig:OC_SVM_Policy_Region}
\end{figure}

OC-SVMs bear similarities to standard Support Vector Machines (SVMs) in that they project training data into a higher dimensional space using a non-linear kernel function, then use a hyper-plane to separate this high-dimensional data. However, OC-SVMs deviate by focusing on a single class and optimizing the hyper-plane to separate the training data from the origin in the hyper-space, thereby creating a non-linear decision function around the data cluster in the lower-dimensional training space. This leads to a binary output from the trained model: a positive number if a data point falls within the defined cluster, expressed by the equation:

\begin{equation}
\label{Eq:OC-SVM}
{\pi}_{Trg}(\textbf{s}_{Trg}) = sgn\left( \sum_{i=1}^{n} \alpha_i K(\textbf{s}_{Trg},\textbf{s}_{i})\right)+\rho
\end{equation}

\noindent or a negative number if the data point is outside the learned cluster. In this equation, $\alpha_i$ denotes the coefficients for the support vectors generated during training, $K$ signifies the Radial Basis Function (RBF) kernel, $\textbf{s}_{Trg}$ refers to the sensory-state vector under test, $\textbf{s}_{i}$ indicates the learned support vectors, and $\rho$ is the bias value of the hyper-plane. The model also incorporates two hand-tuned parameters: the $\gamma$ value within the RBF kernel, affecting the decision boundary's smoothness, and a $\nu$ term that establishes the confidence threshold for data classification within the cluster. Visualization of the learned triggering policy region ($\pi_{Trg}(\textbf{s}_{Trg}) > 0$), the collection of discrete sensory states $\textbf{s}_{Trg}$ separated by a landing rate threshold of 80\%, and a generalized flight trajectory can be seen in Figure \ref{fig:OC_SVM_Policy_Region}.

\subsubsection{Flip Action Policy (Feed-Forward Neural Network)}

\begin{figure}[!t]
    \centering
    \includegraphics[width=1.0\columnwidth]{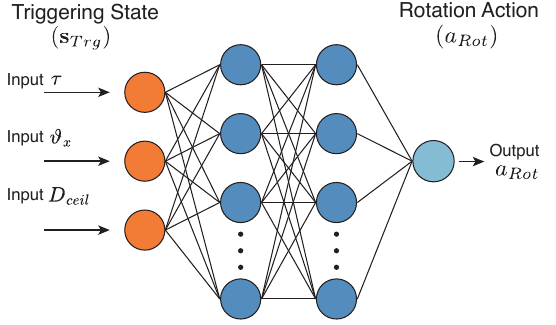}
    \caption{Depiction of feed-forward neural network used to determine body-rotation action ($a_{Rot}$) from the set of triggering sensory states ($\textbf{s}_{Trg}$).}
    \label{fig:Policy_NN}
\end{figure}

Upon triggering, a control action denoted as $a_{Rot}$ is applied to the robot, generating the rotational-maneuver through a self-induced moment about its pitch-axis ($b_y$). This action, $a_{Rot}$, is computed as a function of $\textbf{s}_{Trg}$ according to $a_{Rot}=\pi_{Rot}(\textbf{s}_{Trg})$ \cite{habas2022optimal}. This function serves to generalize the discrete $(\textbf{s}_{Trg},{a_{Rot}})$ pairs into the continuous sensory-space $\mathbb{OF}_{a}$. To determine $\pi_{Rot}$, we utilized supervised machine learning techniques, specifically training a feed-forward neural network, using $(\textbf{s}_{Trg},{a_{Rot}})$ pairs achieving over an 80\% landing success rate as both inputs and outputs. The neural network (Figure \ref{fig:Policy_NN}) features two hidden layers with ten nodes each, optimized for computation speed and size, and utilized the Elu activation function. Throughout the training process, the input data was normalized to zero mean and unit variance, and a Mean-Squared Error loss function was employed to quantify the discrepancy between the training and predicted values.

\section{Method: Simulation and Experimental Setup}

\subsection{Simulation Setup}

Our simulation environment, designed using the Robot Operating System (ROS) and the Gazebo physics environment, was implemented on the Ubuntu 20.04 platform. This environment hosted a quadrotor model based on the Bitcraze Crazyflie 2.1, preserving the physical dimensions and flight parameters of the real-world quadrotor. It also featured a custom Geometric Tracking Controller\cite{lee2013nonlinear} and provided sensory state updates at a frequency of 100 Hz.

The simulation incorporated a ceiling plane as the landing surface, with custom plugins added to model the attachment joint as a ball joint and to record the quadrotor's impact angle upon landing. We also mirrored the mechanical properties of the quadrotor's legs: the hip joint was simulated as a revolute joint and the spring constant and the damping ratio were set at $K = 0.08 \frac{N\cdot mm}{rad}$ and $\zeta = 0.25$ respectively, to emulate the underdamped nature of the physical legs.

Furthermore, our simulation operated at a step size of 0.001 seconds to maintain precise control over the model's behavior. The precision and dependability of our simulation were confirmed in tandem with the system identification steps delineated in Appendix I.

\subsection{Simulation to Real-World Transfer}

\noindent Executing RL in physical systems, especially those characterized by high-speed collisions, presents challenges due to substantial time and monetary investments. Furthermore, the considerable disparity between simulations and real-world outcomes complicates sim-to-real policy transfers. To bridge this gap, our approach consisted of robustly developing a policy in simulation and transferring it to the real-world environment via a zero-shot method. Essential to this was the accurate modeling of the quadcopter in the simulation to enhance the congruence between the simulated and physical models. This was achieved through precise measurement of the quadcopter's rotational inertia, modeling the motor-speed dynamics as a first-order system determined via system identification, and incorporating a battery compensation algorithm in the physical to ensure consistent thrust values throughout battery discharge.

In addition, to bolster policy transfer and resilience, we employed domain randomization during data collection \cite{tobin2017domain}. This involved modulating parameters during learning to enhance the policy's adaptability to environment variations, effectively harnessing diverse simulation data for improved physical performance. Specifically, we randomized inertial parameters, allowing the agent exposure to a spectrum of model variations to avert overfitting. During reinforcement learning data acquisition, both system mass ($m$) and body inertia about the flip-axis ($I_{yy}$) were varied at the onset of each rollout, drawing from Gaussian distributions centered on their base values. With standard deviations set at $\sigma_m = 0.5$ [g] and $\sigma_{Iyy} = 1.5\cdot10^{-6}$ [kg m\textsuperscript{2}], this strategy enhanced simulation diversity and fostered the creation of more robust policies.

\subsection{Landing Gear Designs}

\begin{figure}[!t]
    \centering
    \includegraphics[width=1.0\columnwidth]{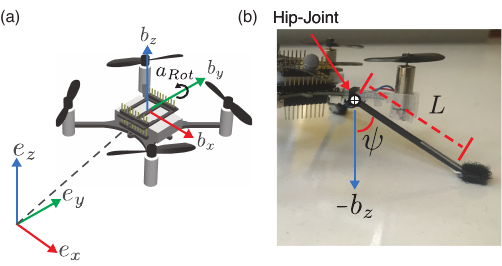}
    \caption{(a) Illustration of the quadrotor's coordinate system relative to the global frame. The body-rotation action ($a_{Rot}$), is realized as a moment about the $b_y$ axis, generated by the front motors.  (b) Diagram depicting the parameters for the leg angle ($\psi$) and leg length ($L$), along with the location of the hinge joint.}
    \label{fig:BodyCoords_LegParams}
\end{figure}

\begin{table}[th]
\caption{Leg Design Configurations}

\begin{center}
        \begin{tabular}{lccc}
                \hline  Leg Design & Angle $\psi$ (deg) & Length $L$ (mm) \\
                \hline 
                \text{ Narrow-Short } & $5^{\circ}$ & 50 \\
                \text{ Narrow-Long }  & $5^{\circ}$ & 75 \\
                \text{ Semi Narrow-Short } & $30^{\circ}$ & 50 \\
                \text{ Semi Narrow-Long } & $30^{\circ}$ & 75  \\
                \text{ Wide-Short } & $60^{\circ}$ & 50 \\
                \text{ Wide-Long } & $60^{\circ}$ & 75 \\
                \hline
        \end{tabular}
\end{center}
\label{table:Leg_Dim}
\end{table}

\noindent Successful dynamic perching largely depends on achieving an appropriate touchdown that brings the robot to a stable landing posture relative to the landing surface. For small aerial robots or animals, this touchdown process can be highly dynamic and involve various forms of soft to hard collisions \cite{Kovac2016}. These landings generally require the robot or animal to have at least one foot firmly planted on the substrate, which is then followed by a rapid body swing or oscillation, culminating in all feet perching on the substrate \cite{habas2022optimal,liu2019flies,Evangelista2010}. Unlike the process of tracking a pre-planned or optimized trajectory, this dynamic touchdown process emerges rapidly from the immediate physical states and properties at the robot-surface interface \cite{Roderick2017}. Therefore, the success of a touchdown process, and consequently the landing, can be significantly influenced by the configurations of the landing gear.

In this work, we explored the influence of various landing gear configurations on inverted landing performance through simulation. We then verified the most effective design via experimental trials where our designs incorporated adhesive feet, capable of adhering to the ceiling, and a hip joint with behavior reminiscent of a torsional spring and damper system (Figure \ref{fig:BodyCoords_LegParams}b). We tested this model experimentally using a VELCRO\textsuperscript{\texttrademark} pad to establish a connection between the landing gear feet and the ceiling. The leg designs were parameterized by their length ($L$) and the angle they formed with the body's $-b_z$ axis ($\psi$) (Figure \ref{fig:BodyCoords_LegParams}b). Our design configurations are detailed in Table \ref{table:Leg_Dim}.

\subsection{Inverted Landing Sequence}

\begin{figure}[!ht]
    \centering
    \includegraphics[width=\linewidth]{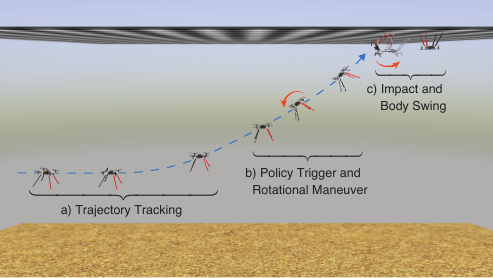}
    \caption{Visualization of our inverted landing sequence, where the quadrotor follows an initial approach trajectory, initiates and executes the landing maneuver, and then impacts and swings into a secure landing position.}
    \label{fig:Flight_Profile}
\end{figure}

\noindent The landing sequence in our inverted landing framework consists of three stages: \textit{1) Trajectory Tracking}: Starting from a hover, the quadcopter follows a pre-defined trajectory, tailored to maintain a specific flight speed and angle according to the desired ceiling-approach conditions. \textit{2) Policy Trigger and Rotational Maneuver}: While progressing along this trajectory, the quadrotor continuously monitors its state in $\mathbb{OF}_{a}$ via emulated sensor data from an external positioning system. Upon meeting the triggering condition defined by $\pi_{Trg}(\textbf{s}_{Trg})$, it executes a rotational maneuver action as dictated by $\pi_{Rot}(\textbf{s}_{Trg})$. \textit{3) Impact and Body Swing}: The robot's fore-legs then make first contact and adhere to the landing surface, leading to a pendulum-like body swing. Landing success is then evaluated based on the robots stabilized landing condition, if there was body/propeller contact, and the number of legs secured to the surface. Depiction of these stages are illustrated in Figure \ref{fig:Flight_Profile}.

\subsection{Physical Experiment Setup}

\noindent In our experiments for physical inverted landing, we utilized the Crazyflie 2.1 nano-sized quadrotor, augmented with four upgraded 19,000 KV brushed DC motors. This compact aerial platform was chosen for its superior maneuverability, supported by an open-source framework, and its enhanced collision resistance due to its small size and the simplicity of part replacement. The quadrotor was equipped with 3D printed legs, designed according to the Semi Narrow-Long configuration (Table \ref{table:Leg_Dim}). For communication, we employed the Crazyswarm ROS package \cite{preiss2017crazyswarm}, establishing a connection between various ROS nodes and the Crazyflie Real Time Protocol embedded in the system's firmware.

As well, we used a Vicon motion capture system to provide real-time position and orientation data, streamed to the Crazyflie at a rate of 100 Hz. This data facilitated the creation of flight trajectories and the simulation of optical flow sensor data, in anticipation of incorporating on-board sensors in future studies.

\begin{figure}[!t]
    \centering
    \includegraphics[width=1.0\columnwidth]{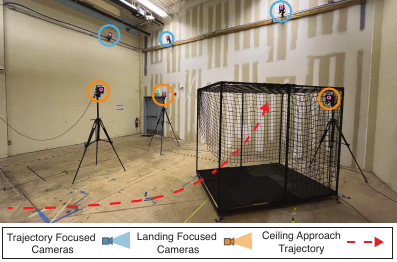}
    \caption{Experimental setup showing an example flight trajectory, the positions of motion capture cameras guiding the flight, and cameras ensuring accurate estimation of triggering state values near the landing surface.}
    \label{fig:Experimental_Setup}
\end{figure}

The experiments were conducted using a portable test cage with dimensions of 1.5m width $\times$ 2.5 m length $\times$ 2.1m height. Velcro pads were fitted onto the ceiling surface to enable the robot's attachment. To capture the flight trajectory in the space near the triggering state, we positioned four Vicon  cameras at the same height as the ceiling landing surface. An additional twelve cameras were positioned above to secure accurate tracking and positioning data throughout flight trajectory prior to triggering (Figure \ref{fig:Experimental_Setup}).

During experimental testing, both $\pi_{Trg}$ and $\pi_{Rot}$, previously trained in the simulation, were uploaded onto the Crazyflie. The quadrotor would begin its flight trajectory outside the test area, accelerating to achieve the predetermined flight speed $\mathbf\|{v}_{ceil}\|$ and flight angle $\mathbf\angle{v}_{ceil}$. This would set it on a collision course with the test surface and prompt it to execute the two-stage landing control policy according to $\pi_{Trg}$ and $\pi_{Rot}$. We designed these flight trajectories to mirror those used in the simulation, allowing us to test various ceiling-approach conditions. The behaviors of each inverted landing trial were recorded, with the process being repeated across a range of feasible flight velocities and angles within our testing environment.

\section{Results and Discussion}

\begin{figure}[!ht]
    \centering
    \includegraphics[width=\linewidth]{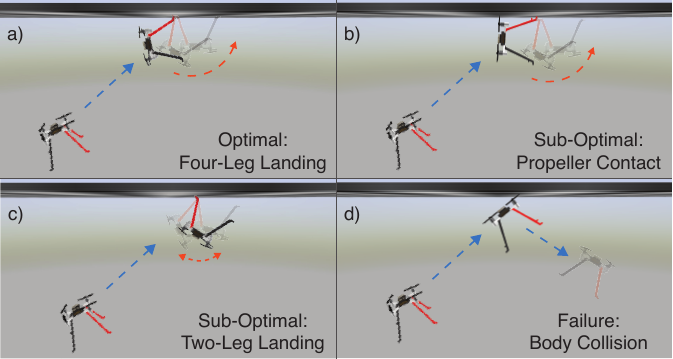}
    \caption{Classifications of Inverted Landing Performance.  (a) Optimal: Successful four-leg landing, no contact with body or propellers. (b) Sub-Optimal: Four-leg landing, but with body or propellers making contact. (c) Sub-Optimal: Incomplete two-leg landing, may include body contact. (d) Failure: Body collision, no leg attachment achieved.}
    \label{fig:Impact_Classifications}
\end{figure}

\subsection{Training Generalized Inverted Landing Control Policy via Optimized Sensory-Action Pairs}

\noindent Utilizing the EPHE algorithm, we focused on collecting optimized sensory-action pairs $(\textbf{s}_{Trg},{a_{Rot}})$ for inverted landing in a simulated environment. During this data collection stage, the magnitude of the ceiling-approach velocity ($\|\mathbf{v}_{ceil}\|$) was incrementally varied from 1.5 to 3.5 m/s, in steps of 0.1 m/s, and the flight angle relative to the horizon ($\mathbf{\angle}\mathbf{v}_{ceil}$) was adjusted from $30^\circ$ to $90^\circ$ in increments of $3.75^\circ$. To enhance the robustness of our following sim-to-real transfer, we also introduced domain randomization by varying the robot's inertial parameters at the beginning of each landing attempt. Also as shown in section V.E, the Semi-Narrow Long leg configuration yielded the best landing performance in our range of experimentally viable test conditions, therefore was used for our initial data collection and policy training/validation work.

The results of this data collection stage demonstrated a range of outcomes in inverted landing success, which we further categorized based on specific characteristics observed during the landings. Here, \textit{Optimal} inverted landings were identified by a distinct sequence: initially, the robot's fore-legs contacted the ceiling, followed by a successful swing that led to the hind-legs touching down, resulting in all four legs adhering to the surface without the body or propellers making contact (see Fig. \ref{fig:Impact_Classifications}a).

A category of \textit{Sub-Optimal} landings were also noted. These landings were often similar to optimal ones, but involved initial contact with either the body or the propellers before stabilizing into a four-leg landing position (Fig. \ref{fig:Impact_Classifications}b). Another sub-optimal scenario involved the robot failing to complete the body swing maneuver, resulting in the hind-legs not contacting the ceiling and only the two fore-legs adhering to the ceiling surface (Fig. \ref{fig:Impact_Classifications}c). This subset included cases with and without propeller or body contact. Finally, the most unsuccessful attempts occurred when the robot collided with the ceiling without any contact between the legs and the landing surface (Fig. \ref{fig:Impact_Classifications}d). For a more comprehensive understanding, detailed visualizations of these landing classifications are available in the accompanying video.

\begin{figure}[!ht]
    \centering
    \includegraphics[width=\linewidth]{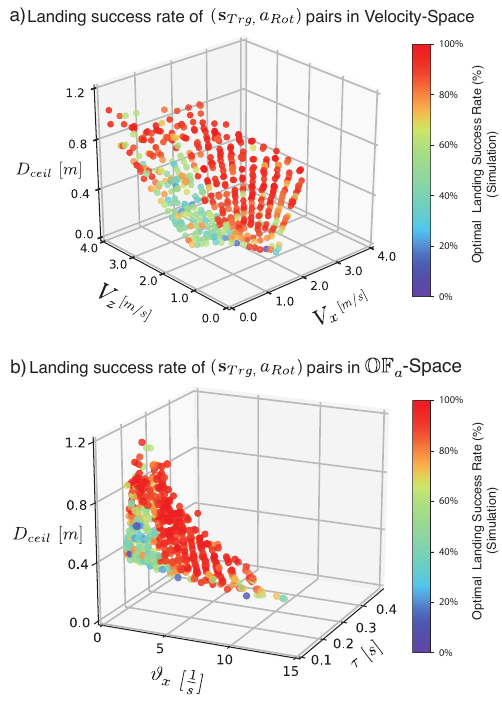}
    \caption{a) Visualization of sensory-action pairs in Velocity-space, highlighting the correlation between higher horizontal velocities and increased inverted landing success in the simulation. b) Visualization of sensory-action pairs in $\mathbb{OF}_{a}$-space, demonstrating a distinct separation between states leading to high and low landing success.}
    \label{fig:Landing_Success_Data}
\end{figure}

Focusing on achieving purely optimal inverted landings (four-leg adhesion with no body/propeller contact), we analyzed the optimal-class landing success rate across various ceiling-approach conditions. All subsequent data will be presented based on this criteria. The collected data for the Semi-Narrow Long configuration is presented in Velocity-space ($V_{x}$, $V_{z}$, $D_{ceil}$) as shown in Fig. \ref{fig:Landing_Success_Data}a. In this data, we observed that a higher velocity magnitude ($\|{v}_{ceil}\|$) generally led to increased success rates and encouraged the triggering of the rotational maneuver further away from the ceiling. Additionally, landings with more angled approaches, indicated by a higher $V_{x}$, yielded higher success rates compared to more vertical approaches, which are characterized by a higher $V_{z}$. For a clearer visualization, the same data is smoothed and presented in Fig. \ref{fig:Policy_Performance_Fig}a, which displays the landing success rates using polar coordinates ($\|\mathbf{v}_{ceil}\|$, $\angle \mathbf{v}_{ceil}$).

\begin{figure}[!ht]
    \centering
    \includegraphics[width=\linewidth]{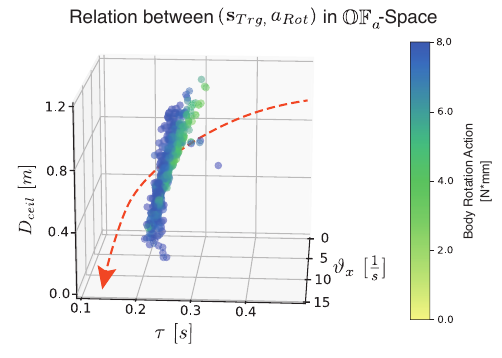}
    \caption{Illustration of the full collection of state-action pairs, where an increase in the rotational moment is observed as the triggering time-to-contact ($\tau_{Trg}$) decreases.}
    \label{fig:Rotation_Action_Data}
\end{figure}

Additionally, the raw dataset of collected ${\textbf{s}}_{Trg}$ values is also depicted in the augmented optical flow space $\mathbb{OF}_{a}$, alongside their corresponding landing success rates (Fig. \ref{fig:Landing_Success_Data}b). In this $\mathbb{OF}_{a}$ space, a clear distinction between high and low landing success rates is evident, allowing for the definition of an enclosed boundary function through setting a success rate threshold (the basis for $\pi_{Trg}$). As well, the relationship between ${\textbf{s}}_{Trg}$ and ${a_{Rot}}$ is illustrated in a side-view via Figure \ref{fig:Rotation_Action_Data}, where the data in $\mathbb{OF}_{a}$ is color-coded based on the magnitude of ${a_{Rot}}$ (the relation that is approximated by $\pi_{Rot}$). This representation highlights a correlation where triggering at lower time-to-contact values values $(\tau)$ is generally associated with higher magnitudes of ${a_{Rot}}$.

In our study, collecting the full dataset for a single leg configuration involved approximately 115,000 simulation-based landing attempts, yielding around 1,000 optimized $(\textbf{s}_{Trg},{a_{Rot}})$ pairs. We established a success criterion for purely optimal landings at an 80\% success rate threshold, leading to the identification of about 300 pairs. These pairs, indicative of optimal four-leg landings without body or propeller contact, formed a distinct cluster in the $\mathbb{OF}_{a}$ space and were crucial for training our generalized two-stage policy. The first stage, the flip triggering policy ($\pi_{Trg}$), generates a boundary function around this subset of training pairs (illustrated by the shaded area in Fig. \ref{fig:OC_SVM_Policy_Region}), dictating the precise moment to initiate the rotational maneuver. The second stage, the rotational maneuver policy ($\pi_{Rot}$), then calculates the necessary body moment based on the triggering state.

\subsection{Validation of the Generalized Inverted Landing Policy in Simulation}

\begin{figure}[!t]
    \centering
    \includegraphics[width=1.0\columnwidth]{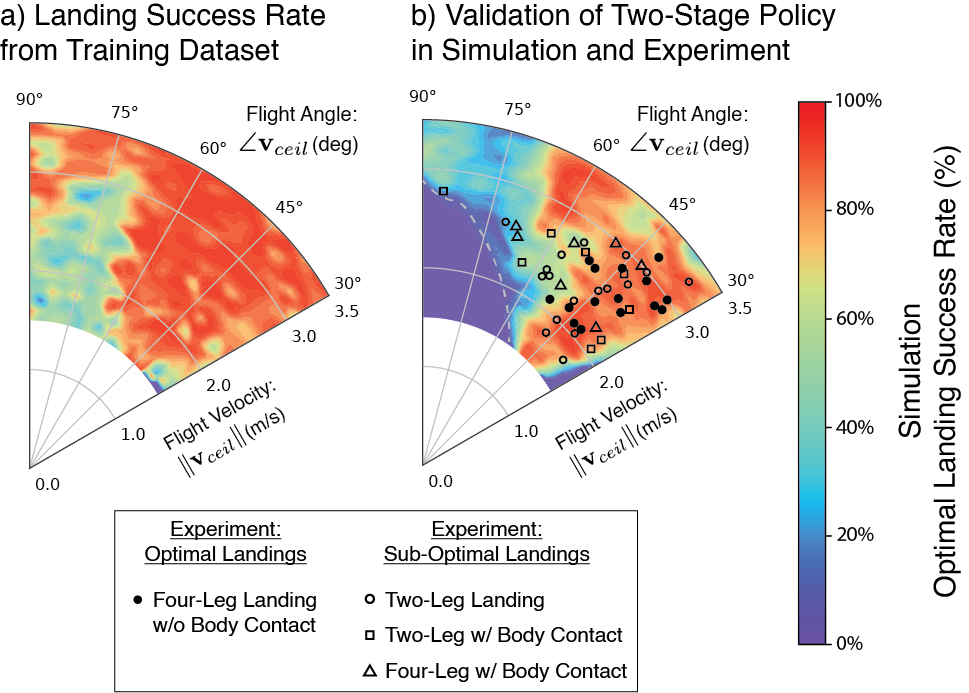}
    \caption{(a) Polar plot showcasing the overall inverted landing capabilities derived from the simulated training dataset specific to the Narrow-Long leg configuration. (b) A comparison of simulation-based results with experimental outcomes for our formulated two-stage policy.
    }
   \label{fig:Policy_Performance_Fig}
\end{figure}

\noindent Upon completion of training our generalized two-stage inverted landing control policy, we conducted validation tests in both simulated and experimental environments. In the simulation, the policy was tested under flight conditions similar to those used for collecting the initial sensory-action pairs. The results of which indicated that at lower ceiling-approach angles (between $30^\circ$ and $65^\circ$), our generalized control policy (Fig. \ref{fig:Policy_Performance_Fig}b) performed comparably to the optimal discrete sensory-action pairs (Fig. \ref{fig:Policy_Performance_Fig}a). However, in near-vertical approach conditions (from $65^\circ$ to $90^\circ$), we observed moderate performance degradation. This result was expected due to the prevalence of successful sensory-action pairs at lower angles compared to the more vertical approaches, as shown in Figure \ref{fig:Landing_Success_Data}a.
Consequently, this led to a training bias in the two-stage control policy towards lower angle approaches, further illustrated in background plot of Fig. \ref{fig:Policy_Performance_Fig}b.

\begin{figure*}[!t]
    \centering
    \includegraphics[width=1.0\textwidth]{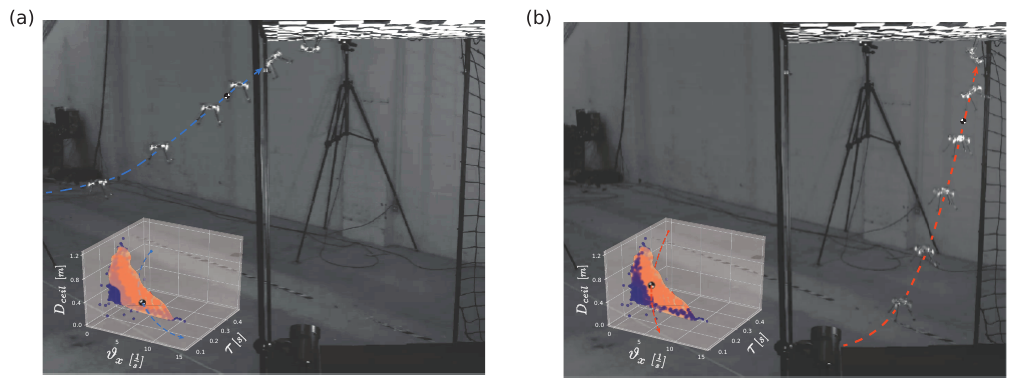}
    \caption{(a) Experimental inverted landing sequence for the flight condition ($2.50$ m/s, $37^\circ$) resulting in a four-leg successful landing. The trajectory through $\mathbb{OF}_{a}$ space is depicted, with emphasis on the intersection point with the boundary region in both Cartesian and $\mathbb{OF}_{a}$ spaces. (b) A corresponding depiction for the flight condition ($2.60$ m/s, $70^\circ$), leading to a two-leg landing.}
    \label{fig:Exp_Flight_Traj}
\end{figure*}

\subsection{Validation of the Two-Stage Inverted Landing Policy in Experiments}

\noindent In transitioning from simulation to experimental application and to validate the trained two-stage policy, we employed a zero-shot sim-to-real transfer approach. This validation was conducted using a Bitcraze Crazyflie 2.1 quadrotor, the same model our simulation was based on. For experimental testing, we selected ceiling-approach conditions with lower flight angles, primarily in the range of 30 to 65 degrees, which correlated with higher landing success rates in simulation and where our control policy demonstrated comparable performance to the collected sensory-action pairs. It is important to note that due to limitations such as the robot's thrust capacity and the available space, achieving high vertical velocity approaches were difficult to achieve in our experiments. However, these conditions typically resulted in lower success rates in simulation and were not chosen for experimental validation.

The results of our testing revealed a mixture of both optimal (i.e., characterized by four-leg touchdown and no body/propeller contact) and sub-optimal (i.e., two-leg landings or four-leg landings with body/propeller contact) landing outcomes; as illustrated in Figure \ref{fig:Policy_Performance_Fig}b. Notably, the simulation-based data using in training our control policy consisted primarily of optimal-class landings outcomes, so the prevalence of sub-optimal landings indicate degradation of our two-stage policy when transferring from our simulation-based environment to our physical experiment environment.

To this end, optimal landings were predominantly observed at flight angles between 30 and 45 degrees; an example of such a landing is illustrated in Fig. \ref{fig:Exp_Flight_Traj}a. Conversely, sub-optimal landings were present across all tested conditions, often overlapping with optimal landings without a distinct separation. Figure \ref{fig:Exp_Flight_Traj}b shows an example of such a sub-optimal landing. Notably, all flight conditions intersecting the triggering boundary function ($\pi_{Trg}$) achieved some level of success, resulting in either optimal or sub-optimal landings, with no direct failures recorded.
Flight conditions outside the boundary function limits, represented in Fig. \ref{fig:Policy_Performance_Fig}b as the dark purple area inside the grey-dashed line, were not tested as they fell beyond the feasible range of our control policy, posing a high risk of damage to the robot.

These experimental tests showed reduced performance compared to our simulation results, likely due to several factors. A major discrepancy stemmed from inaccuracies in the motor model, affecting both flight dynamics and body-rotation characteristics, and in the simulation of the landing gear/hinge joint. Specifically, the current model fails to fully capture the deformation behavior of the 3D printed landing gear. To overcome these issues, future work should expand use of domain randomization to encompass a wider range of system variabilities. Additionally, improving model fidelity and refining the reward function to emphasize robustness could significantly increase the success rate of experimental inverted landings

\begin{figure*}[!t]
    \centering
    \includegraphics[width=1.0\textwidth]{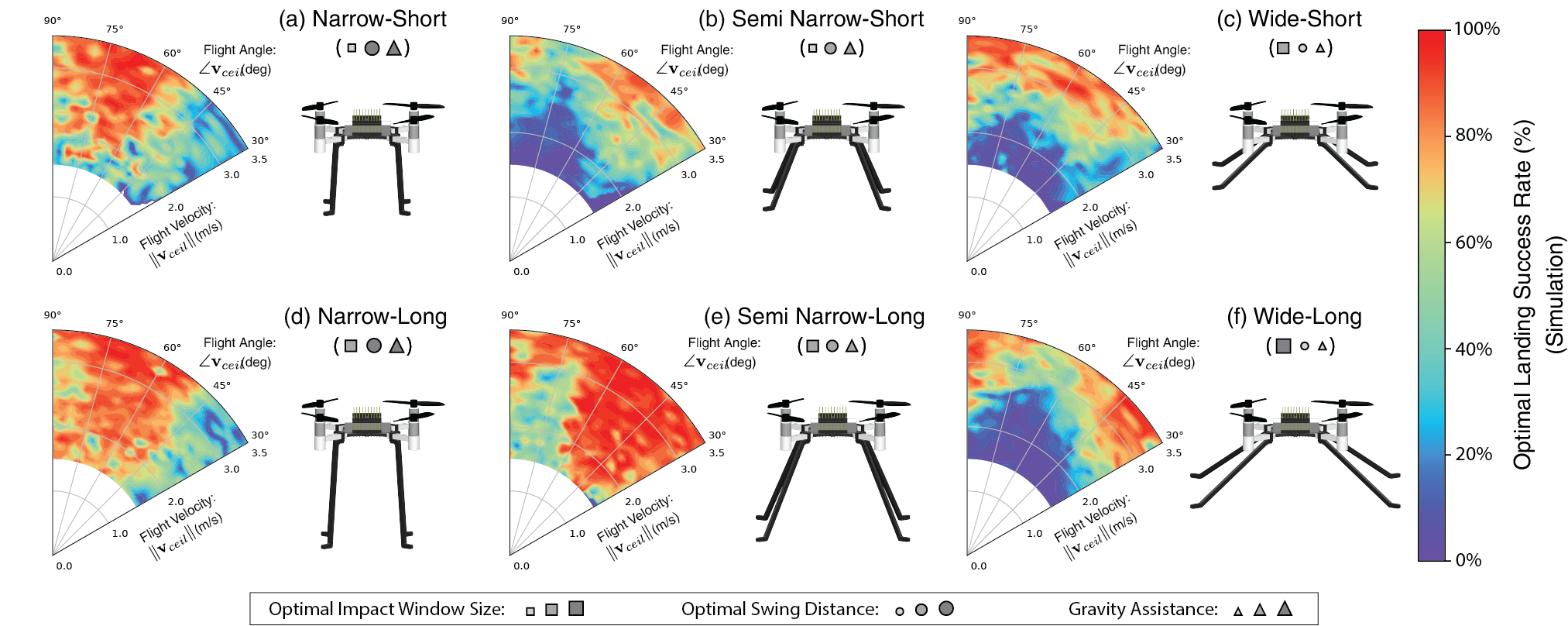}
    \caption{Polar plots indicating the landing success rate found in simulation for various leg design configurations. The radial axis corresponds to the flight velocity magnitude ($\|\mathbf{v}_{ceil}\|$), while the angular axis represents the flight angle ($\angle \mathbf{v}_{ceil}$). Data was gathered for parameters where $\|\mathbf{v}_{ceil}\|$ ranges from 1.5 to 3.5 m/s and $\angle \mathbf{v}_{ceil}$ spans from $30^{\circ}$ to $90^{\circ}$.}
    \label{fig:Polar_Plots}
\end{figure*}

\subsection{Impact of Leg Design and Flight Conditions on Quadrotor Inverted Landing Efficacy}

Finally, we simulated various leg configurations and investigated their impact on landing success across flight conditions; the details of the configurations tested are presented in Table \ref{table:Leg_Dim}. Furthermore, for data collection we applied the same parameter optimization approach previously described, focusing on a range of flight velocity magnitudes ($\|\mathbf{v}_{ceil}\|$) and directions ($\angle \mathbf{v}_{ceil}$) consistent with our initial data collection methodology shown in Section V.A.

Our study revealed significant variations in landing success based on the leg configuration. Specifically, the Semi-Narrow configurations demonstrated higher success at lower flight angles (30 deg to 65 deg), as illustrated in Figure \ref{fig:Polar_Plots}b,e. However, their effectiveness declined markedly in near-vertical flight conditions (above 65 deg). In contrast, the Narrow configurations excelled in more vertical flights (above 50 deg) but were less successful at angles below 50 deg, as shown in Figure \ref{fig:Polar_Plots}a,d. The Wide configurations generally under-performed in comparison, except at very low or extremely vertical angles (Figure \ref{fig:Polar_Plots}c,f). Across these configurations, we also observed a trend where higher flight velocity magnitudes correlated with increased inverted landing success. Additionally, shorter leg lengths generally exhibited less success across all flight conditions compared to their longer counterparts, as depicted in Figure \ref{fig:Polar_Plots}a-c.

The diversity in quadrotor inverted landing behaviors can primarily be attributed to the balance of four key factors: 1) \textit{Swing Distance about Fore-leg Contact}; 2) \textit{Gravitational Contribution to the Swing}; 3) \textit{Sufficiency of Momentum Transfer}; 4) \textit{Impact Window Size}. These factors are significantly influenced by the body angle at impact, the quadrotor's leg geometry, and the flight conditions when the landing policy is triggered. This complex interaction leads to varying inverted landing capabilities across different configurations

\begin{figure}[!t]
    \centering
    \includegraphics[width=1.0\columnwidth]{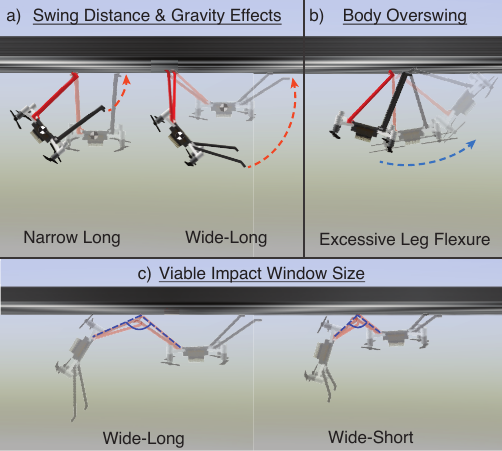}
    \caption{a) For the same body orientation at impact, quadrotors with narrower leg designs require a shorter swing distance to make contact with the landing surface. b) Leg joints with excessive flexure can lead to overswing and body contact with the landing surface. c) Shorter-legged designs typically lead to a narrower range of viable impact conditions compared to their longer leg counterparts.}
    \label{fig:Landing_Dynamics}
\end{figure}

Viewing through this framework lens, it can be seen that minimizing the \textit{Swing Distance about Fore-leg Contact} significantly enhances inverted landing success. Here the impact angle and the quadrotor's leg geometry directly determine the necessary swing distance for the hind legs to make contact with the surface. Figure \ref{fig:Landing_Dynamics}a illustrates this, where, for a given body angle at impact, the Narrow-Long geometry necessitates a smaller swing than the Wide-Long configuration to achieve the desired landing state, thereby reducing the travel distance and energy required.

Furthermore, maximizing the \textit{Gravitational Contribution to the Swing} significantly enhances landing success. This effect is influenced by the quadrotor's impact angle and leg geometry, which can result in the Center of Mass (CoM) being lower at the final state than at the time of impact. Figure \ref{fig:Landing_Dynamics}a illustrates this for the Narrow-Long leg configuration, where gravity aids in achieving the desired state with minimal effort. Conversely, configurations like the Wide-Long, also shown in Figure \ref{fig:Landing_Dynamics}a, may require the CoM to be higher in the final state than at impact, necessitating additional energy for a successful body-swing.

Balancing the \textit{Sufficiency of Momentum Transfer} is crucial for inverted landing success, even when other factors are unfavorable. This aspect depends on the body angle at impact, leg geometry, and the quadrotor's flight conditions. Here, properly aligning the impact angle to convert translational momentum into rotational momentum allows the robot to cover the necessary swing distance and overcome any adverse gravitational effects. Failure to adequately coordinate this can lead to a two-leg landing due to insufficient rotational momentum. Additionally, some combinations of flight conditions and leg geometry may not provide enough translational momentum for a successful four-leg inverted landing. Our supplemental video further illustrates the importance of momentum management, swing distance, and gravitational effects for achieving inverted landing success

Additionally, the leg configuration's geometry directly affects the \textit{Impact Window Size}, determining the range of viable impact angles that ensure only the legs, and not the body, contact the ceiling, as shown in Figure \ref{fig:Landing_Dynamics}c. Shorter-legged designs (compared to their longer counterparts) typically offer a narrower range of viable impact windows, increasing the need for precision in impact conditions. In systems with significant noise, like ours, this increased sensitivity can significantly affect the robustness and success rate of inverted landings.

The direct effects of these concepts are evident in the results presented and in Figure \ref{fig:Polar_Plots}. For example, the Semi-Narrow configurations (Figure \ref{fig:Polar_Plots}b,e) are notably effective under lower flight angles, and capable of striking a balance between manageable swing distances, gravitational contributions, and sufficient momentum transfer, unlike when under more vertical flight conditions. On the other hand, the Narrow configurations (Figure \ref{fig:Polar_Plots}a,d) perform well across a wide range of vertical flight conditions, benefiting from their design, which results in smaller swing distances and gravitational assistance. Their reduced success at lower flight angles, however, was attributed to excessive leg flexure leading to an overswing behavior and propeller contact, as depicted in Figure \ref{fig:Landing_Dynamics}b. This issue, linked to the stiffness of the leg joints, will be a focus of future research, with the anticipation that stiffer joints could enhance performance at lower flight angles.

Moreover, the Wide configurations, with their significantly larger swing distances due to leg geometry, and detrimental gravitational effects, demand higher translational velocities in flight for successful landings. While sufficient momentum transfer can be achieved in very low or extremely vertical flight angles, it becomes challenging in intermediate cases. Identifying and modelling the specific aspects which determine if a sufficient balance between swing distance, gravitational effects, and efficient momentum transfer factors can be achieved will also be the topic of future research. Finally, across all leg angles (i.e., Narrow, Semi-Narrow, Wide), the shorter leg configurations have a smaller \textit{Impact Window Size}, which led to reduced performance across flight conditions (Figure \ref{fig:Landing_Dynamics}a-c),  compared to their longer leg counterparts (Figure \ref{fig:Landing_Dynamics}d-f). 

\section{Conclusion and Future Work}
In this study, we aimed at achieving inverted landing in robotic fliers and explored the concept of dynamic perching, a skill mastered by many species birds, bees, and bats, yet a significant challenge in robotics. Our methodology centers on a biologically inspired, two-stage control policy that reflects the adept sensory-motor mapping observed previously in flies and utilizes augmented optical flow cues. It is formulated from sensory-action pairings identified through parameter optimization in simulation and further refined using various machine learning techniques. The policy translates emulated augmented optical-flow data into motor control actions in terms of initiating (in the first stage) and modulating (in the second stage) the rotational maneuvers that are critical for successful inverted landing. Our approach, distinct from traditional full-state trajectory-based methods, promotes a range of emergent behaviors through computationally-efficient reactive motor control. The effectiveness of this comprehensive policy, proven in both simulated and real-world environments, highlights its potential in advancing dynamic robotic perching. 

Additionally, our study of varying leg geometries highlights the complex relationship between design, control, and function, emphasizing the importance of mechanical intelligence. Through testing various configurations, it is clear that leg geometry significantly influences the success and robustness of inverted landings. Factors like swing distance around fore-leg contact, and efficiency of momentum transfer at impact, which are crucial in determining inverted landing success, are directly influenced by the parameters of leg design. In our results, we found that narrower leg configurations lead to shorter swing distances, enhancing inverted landing robustness, particularly in vertical flight scenarios compared to more horizontal flight. As well, achieving a sufficiently high velocity is essential to bolster landing capabilities and ensure effective contact. 

In conclusion, our research represents significant progress in emulating the perching capabilities of nature's most adept fliers. The integration of computational and mechanical intelligence is crucial in this pursuit. As the field of aerial robotics advances, emulating the perching expertise of birds and insects is becoming more attainable. Future work will aim to 1) directly estimate the  sensory inputs from onboard cameras and IMUs and 2)  to refine our two-stage control methodology and enhance the synergy between computational strategies and physical design. By deepening our understanding of these elements and unraveling their interdependence, we will move closer to achieving the autonomous robust perching capabilities in small aerial robots that will enable them to fly reliably and safely in cluttered urban or natural environments.

\section*{Acknowledgments}
We would like to extend our sincere gratitude to Zafar Anwar for his invaluable assistance with slow-motion video recording, and the OpenAI GPT-4 model for contributing to the editing process of this manuscript. This research was made possible by the National Science Foundation grants IIS-1815519 and CMMI-1554429 awarded to B.C., as well as the support of the Department of Defense (DoD) through the National Defense Science \& Engineering Graduate (NDSEG) Fellowship Program awarded to B.H.. We would like to express our appreciation to these individuals and organizations for their support of our research efforts.

{
\appendices
\section{System Identification}

\subsection{Inertia Estimation}

Given our reliance on a zero-shot sim-to-real transfer method, the success of our two-stage control policy and the execution of inverted landings are strongly tied to the simulation's accuracy, thus emphasizing the need for precise rotational inertia estimation. To this end, we applied the bifillar pendulum method \cite{garcia2017modeling,jardin2009optimized} to measure the rotational inertia about the quadrotor's three principal axes. The technique involved suspending the quadrotor using two vertical strings and introducing a minor rotational displacement, thereby inducing oscillation about the vertical axis, as illustrated in Figure \ref{fig:Bifillar_Pendulum}. Using the onboard gyro sensor, we measured the oscillation period and calculated the average time between peaks. The moment of inertia was then estimated using the following equation:

\begin{equation}
\label{Eq:Inertia_Estimation}
I_{est} = \frac{m g \ (D \ T_{avg})^2}{L \ (4 \pi)^2}  ,
\end{equation}

where $mg$ is the weight of the quadrotor in Newtons, $D$ is the distance between the strings, $L$ is the string length, and $T_{avg}$ is the average oscillation period. We repeated this procedure for each axis of the system. The results for the standard Crazyflie, as well as our modified system with attached legs, are presented in Table \ref{table:Inertia_Estimates}.

\begin{table}[ht]
\renewcommand{\arraystretch}{1.3} % adjust the value as you want
\caption{Measured Body Moment of Inertia}
\begin{center}
\begin{tabular}{lcc}
\hline
& Crazyflie 2.1 - Stock & Crazyflie 2.1 - NL Legs \\ \hline
Mass [g] & 30.0 & 38.1 \\
$I_{xx}$ [10$^{-6}$ kg m$^{2}$] & 12.19 & 27.93 \\ 
$I_{yy}$ [10$^{-6}$ kg m$^{2}$] & 14.55 & 30.46 \\
$I_{zz}$ [10$^{-6}$ kg m$^{2}$] & 23.55 & 47.12 \\ \hline
\end{tabular}
\end{center}
\label{table:Inertia_Estimates}
\end{table}

\begin{figure}[!t]
    \centering
    \includegraphics[width=0.9\columnwidth]{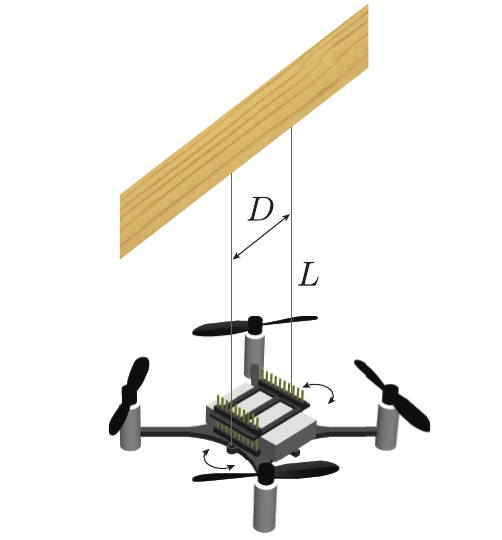}
    \caption{Bifillar pendulum setup used to estimate rotational inertia about the quadrotor's primary axes. $D$ represents the distance between mounting points and $L$ represents the length of the string.}
    \label{fig:Bifillar_Pendulum}
\end{figure}

\subsubsection{Thrust-Battery Compensation}

Maintaining the desired thrust values from the quadrotor's motors throughout the entire flight is essential for the accuracy and consistency of our experiments. This includes ensuring the thrusts used in experimental settings align with those in our simulation. To address this, we introduced a battery compensation algorithm that guarantees consistent thrust values through PWM modulation---irrespective of fluctuations in battery voltage.

The battery compensation algorithm we developed, an enhancement of Bitcraze's open-source framework, is specifically designed to provide stable thrust during flight, counterbalancing the effects of battery depletion. Traditional brushed DC motors, like the ones originally equipped on the Crazyflie, generate thrust by applying a fraction of the available voltage to the motors via a PWM signal. However, as the battery voltage decreases over time, a constant PWM signal can lead to thrust inconsistency. To mitigate this, our algorithm modulates the PWM signal, maintaining a consistent voltage applied to the motors despite decreasing battery voltage, thereby ensuring consistent thrust.

To address fluctuations in the Crazyflie's battery voltage affecting motor performance, we established a regression-based approach to discern the voltage required for given thrust commands. Using a predictive curve, we modeled the voltage-thrust relationship, translating controller thrust values to the appropriate motor voltage.

To derive this curve, the Crazyflie quadrotor was powered by an external supply at a fixed voltage and to negate ground effect interference, the quadrotor was suspended over an airspace and anchored to a scale. We then systematically varied PWM values, logging the consequent thrust, supply voltage, and measured onboard voltage. The motor voltage, calculated as $V_{motor} = V_{onboard} \cdot \frac{PWM_{cmd}}{PWM_{Max}}$, was then plotted against the measured thrusts. The resultant data was then fitted to a logarithmic curve, described by the equation:

\begin{equation}
\label{Eq:Thrust_Voltage_Eq}
V_{motor} = a \cdot ln(f_{thrust} - b) + c .
\end{equation}

For improved precision, especially at higher thrust values, we bifurcated the curve into two regions (refer to Figure \ref{fig:Batt_Comp_Curves}a). In practice, the resultant PWM command is computed from the current battery voltage, desired voltage, and the requested motor thrust. This computation ensures constant thrust throughout the flight.

The efficacy of our algorithm is validated in Figure \ref{fig:Batt_Comp_Curves}b. Here, we contrast the thrust values generated by our battery compensation and PWM modulation algorithm with those resulting from a constant PWM value over the life of a charged batttery. Both the standard Crazyflie 2.1 motors and our upgraded BetaFPV 7x16mm 19,000 KV motors underwent this procedure. The parameters identified through this process are detailed in Table \ref{table:BatteryComp_Params}.

\begin{figure}[!t]
    \centering
    \includegraphics[width=1.0\columnwidth]{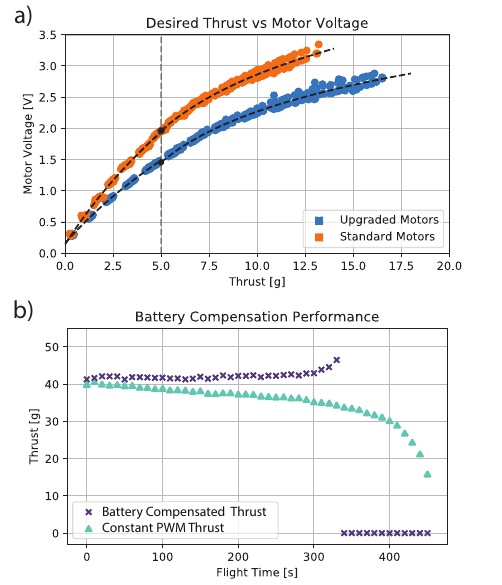}
    \caption{(a) Plot illustrating the relationship between desired motor thrust and the necessary motor voltage ($V_{motor}$) for both standard and upgraded motors. (b) Plot depicting the efficacy of our upgraded motors using the battery compensation algorithm in comparison to a constant PWM strategy.}
    \label{fig:Batt_Comp_Curves}
\end{figure}

\begin{table}[!h]
    \caption{Battery Compensation Parameters}
    \renewcommand{\arraystretch}{1.3} % adjust the value as you want
    \begin{center}
        \begin{tabular}{lccc}
            \hline
            & \multicolumn{3}{c}{$V_{motor} = a \cdot ln(b \cdot f_{thrust}) + c $} \\
            \hline
            Thrust $\leq$ 5g & $a$ & $b$ & $c$ \\
            \hline 
            Stock Motors & 0.618 & 1.394 & 0 \\
            Upgraded Motors & 1.285 & 1.512 & 0 \\
            \hline 
            Thrust $>$ 5g & $a$ & $b$ & $c$ \\
            \hline 
            Stock Motors & 2.097 & -4.464 & 5.636 \\
            Upgraded Motors & 3.230 & -5.469 & 5.979 \\
            \hline
        \end{tabular}
    \end{center}
    \label{table:BatteryComp_Params}
\end{table}

\subsubsection{Motor-Speed Dynamics}
To further enhance the precision of our simulation, we incorporated rotor acceleration dynamics by modeling the motor thrusts as a first-order system. This is an improvement on previous work by the authors which assumed an instantaneous motor response to thrust \cite{habas2022optimal}, neglecting the decay behavior of rear motors that relies solely on air drag forces for slowing down. The inclusion of this behavior in our model greatly improved the accuracy between our simulation and physical experiments.

For a precise representation of motor behavior, we established time-constants ($\tau_{up}/\tau_{down}$) to model the first-order system's speed-up and slow-down dynamics. To do this we built a custom tachometer, comprising an IR detection sensor and an Arduino micro-controller, to record the motor's time profile for varying speed changes. Using the conversion term from Forster et al. \cite{forster2015system}, we obtained the corresponding thrust values from propeller speeds. Then, an exponential curve was fitted to these profiles (Figure \ref{fig:Motor_Thrust_Decay}), yielding the speed-up and slow-down time constants ($\tau_{up} = 0.05$ and $\tau_{down} = 0.16$).

\begin{figure}[!thbp]
    \centering
    \includegraphics[width=1.0\columnwidth]{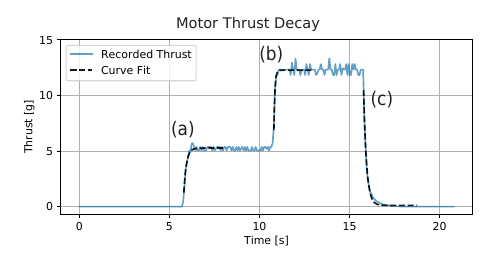}
    \caption{Figure depicting the quadrotor motor's acceleration and deceleration characteristics, modeled as a first-order system. The curve fit time constants are: (a) $\tau_{up} = 0.06s$; (b) $\tau_{up} = 0.05s$; (c) $\tau_{down} = 0.16s$ .}
    \label{fig:Motor_Thrust_Decay}
\end{figure}
}

\bibliography{IEEEabrv, refs}
\bibliographystyle{IEEEtran}

\newpage

\begin{IEEEbiography}[{\includegraphics[width=1in,height=1.25in,clip,keepaspectratio]{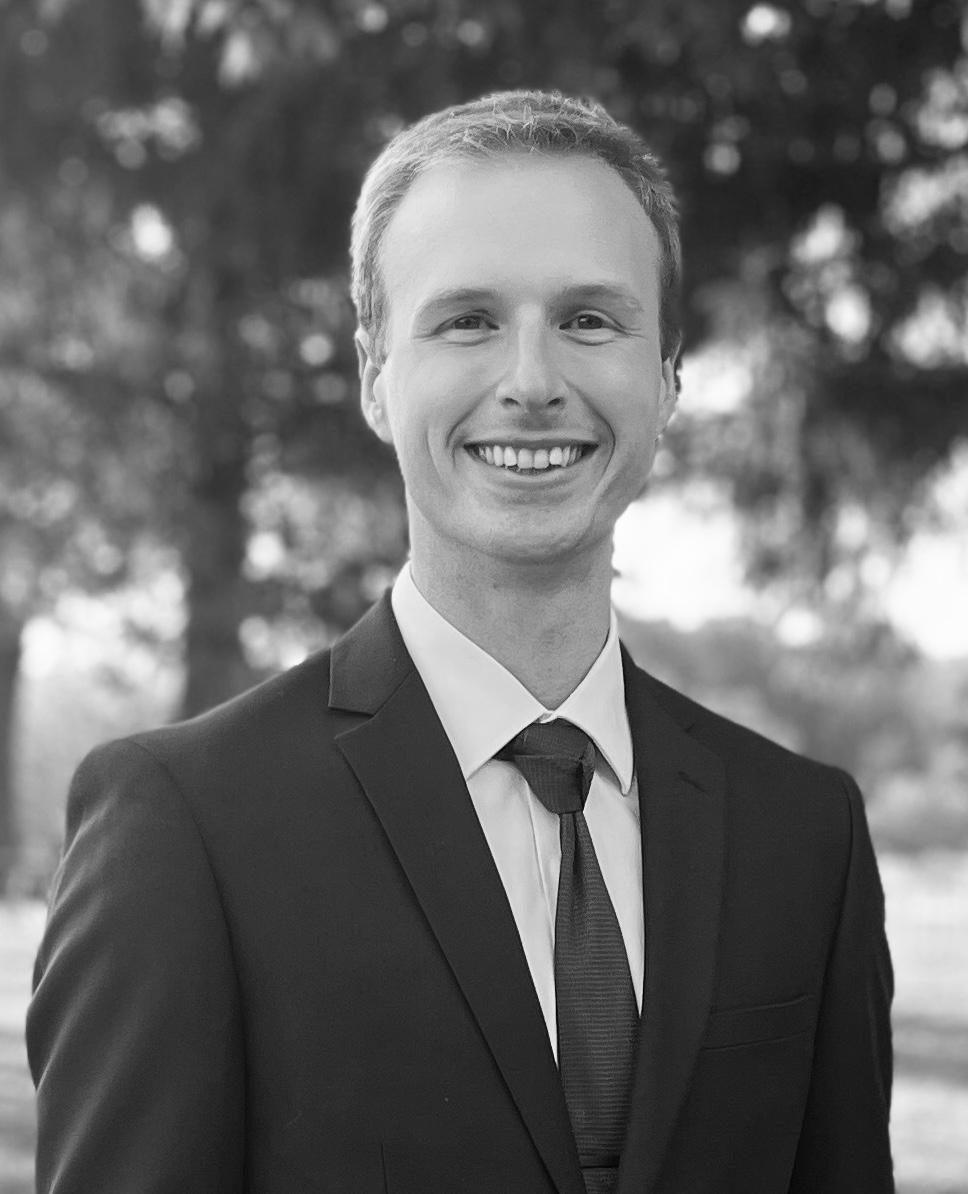}}]{Bryan Habas} is currently pursuing his Ph.D. at The Pennsylvania State University, under the mentorship of Professor Cheng in the BioRob-InFL laboratory. He earned his Master of Science in Mechanical Engineering from the same institution in 2021. Bryan’s research endeavors encompass a broad spectrum of areas including aerial robotics, control systems, applied machine learning, and the simulation of dynamic systems.
\end{IEEEbiography}

\begin{IEEEbiography}[{\includegraphics[width=1in,height=1.25in,clip,keepaspectratio]{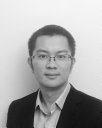}}]{Bo Cheng (Member, IEEE)} received the B.S. degree in automation from Zhejiang University, Hangzhou, China, in 2006, the M.S. degree in mechanical engineering from the University of Delaware, Newark, DE, USA, in 2009, and the Ph.D. degree in mechanical engineering from Purdue University, West Lafayette, IN, USA, in 2012. He is currently an Associate Professor with the Department of Mechanical Engineering, Pennsylvania State University, University Park, PA, USA. His research interests include bioinspired robotics, micro aerial vehicles, robot control and learning, insect and hummingbird flight, and bioinspired fluid dynamics.

\end{IEEEbiography}

\vfill

\end{document}